\pdfoutput=1

\documentclass[11pt]{article}

\usepackage{EMNLP2023}  

\usepackage{times}
\usepackage{latexsym}

\usepackage[T1]{fontenc}

\usepackage[utf8]{inputenc}

\usepackage{microtype}

\usepackage{inconsolata}

\title{nerblackbox: \\ A High-level Library for Named Entity Recognition in Python}

\author{Felix Stollenwerk \\
  AI Sweden \\
  \texttt{felix.stollenwerk@ai.se} \\
  }

\usepackage{lipsum}    
\usepackage{comment}   
\usepackage{graphicx}  
\usepackage{pifont}    
\usepackage{float}     
\usepackage{booktabs}  
\usepackage{listings}
\usepackage{xcolor}    
\definecolor{codegreen}{rgb}{0,0.6,0}
\definecolor{codegray}{rgb}{0.5,0.5,0.5}
\definecolor{codepurple}{rgb}{0.58,0,0.82}
\definecolor{backcolour}{rgb}{0.95,0.95,0.92}
\lstdefinestyle{mystyle}{
    backgroundcolor=\color{backcolour},   
    commentstyle=\color{codegreen},
    keywordstyle=\color{magenta},
    numberstyle=\tiny\color{codegray},
    stringstyle=\color{codepurple},
    basicstyle=\ttfamily\footnotesize,
    breakatwhitespace=false,         
    breaklines=true,                 
    captionpos=b,                    
    keepspaces=true,                 
    numbers=left,                    
    numbersep=5pt,                  
    showspaces=false,                
    showstringspaces=false,
    showtabs=false,                  
    tabsize=2
}
\newcommand{\code}[1]{\colorbox{backcolour}{\ttfamily\footnotesize #1}}

\newcommand{\X}{\ding{53}}

\newcommand{\nerblackbox}{\textit{nerblackbox}}
\newcommand{\transformers}{\textit{transformers} \cite{wolf_transformers_2020}}
\newcommand{\transformersNO}{\textit{transformers}}
\newcommand{\spacy}{\textit{spacy} \cite{honnibal_spacy_2020}}
\newcommand{\datasets}{\textit{datasets} \cite{lhoest_datasets_2021}}
\newcommand{\evaluate}{\textit{evaluate} \cite{von_werra_evaluate_2022}}
\newcommand{\seqeval}{\textit{seqeval} \cite{nakayama_seqeval_2018}}

\newcommand{\mlflow}{\textit{MLflow}\footnote{\url{https://pypi.org/project/mlflow/}}}
\newcommand{\tner}{\textit{T-NER} \cite{ushio_t-ner_2021}}
\newcommand{\tnerNO}{\textit{T-NER}}
\newcommand{\simpletransformers}{\textit{Simple Transformers} \cite{rajapakse_simple_2019}}
\newcommand{\simpletransformersNO}{\textit{Simple Transformers}}
\newcommand{\labelstudio}{\textit{LabelStudio} \cite{tkachenko_label_2020}}
\newcommand{\doccano}{\textit{Doccano} \cite{nakayama_doccano_2018}}
\newcommand{\mypy}{\textit{mypy}\footnote{\url{https://pypi.org/project/mypy/}}}
\newcommand{\black}{\textit{black}\footnote{\url{https://pypi.org/project/black/}}}

\begin{document}
\maketitle
\begin{abstract}
We present \nerblackbox, a python library to facilitate the use of state-of-the-art transformer-based models for named entity recognition. It provides simple-to-use yet powerful methods to access data and models from a wide range of sources, for fully automated model training and evaluation as well as versatile model inference. While many technical challenges are solved and hidden from the user by default, \nerblackbox~also offers fine-grained control and a rich set of customizable features. It is thus targeted both at application-oriented developers as well as machine learning experts and researchers. 
\end{abstract}

\section{Introduction}

\begin{figure}[!ht]
    \centering
    \includegraphics[scale=0.38, trim=50 0 40 0]{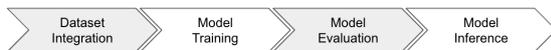}
    \caption{Essential stages in the life cycle of a machine learning model.}
    \label{fig:workflow}
\end{figure}

Named Entity Recognition (NER) is an important natural language processing task with a multitude of applications \cite{lorica_2021_2021}. 
While generative AI is currently ubiquitous in the scientific literature and public debate, it has not (yet) replaced discriminative AI for information extraction tasks like NER. Fine-tuned, transformer-based encoder models are both SOTA in research\footnote{\url{http://nlpprogress.com/english/named_entity_recognition.html}}
and commonly used by developers to solve real-world problems, see e.g.~\cite{raza_large-scale_2022, stollenwerk_annotated_2022}.
Popular open source frameworks, like the ones provided by HuggingFace \cite{wolf_transformers_2020, lhoest_datasets_2021, von_werra_evaluate_2022}, greatly facilitate the use of such models. 
They cover the whole workflow consisting of dataset integration, model training, evaluation and inference, see Fig.~\ref{fig:workflow}.

However, they do require a certain degree of expertise and often some significant, use-case specific effort. 
Some of the (general and NER-specific) challenges are: 

\paragraph{(i)}
There exist various sources for datasets. Regarding public datasets, HuggingFace and GitHub repositories are important sources. Private datasets may be stored on local filesystems or be created using annotation tools. 
Additional complexity is introduced by the circumstance that datasets often come in different formats. This may be true even for datasets from the same source. 
These issues typically require customized data preprocessing code for every new use case.  

\paragraph{(ii)}
Data for NER is processed on three different levels: tokens, words and entities. Different parts of the workflow may operate on different levels, as shown in Tab.~\ref{tab:levels}. 
\begin{table}[!ht]
    \centering
    \small
    \begin{tabular}{lccc}
         \toprule
         \textbf{stage} & \textbf{token} & \textbf{word} & \textbf{entity} \\ 
         \midrule
         dataset & & \X & \X \\
         training & \X & \X & \\
         evaluation & & & \X \\
         inference & & \X & \X
    \end{tabular}
    \caption{Overview of the data levels that the different parts of a NER model workflow can operate on.}
    \label{tab:levels}
\end{table}
Datasets may be pretokenized (word level) or not (entity level). At training time, labels for tokens that are not the first token of a word may be ignored (word level) or included (token level) in the computation of the loss. Model evaluation takes place primarily on the entity level (although it is labels on the token or word level that are employed for the computation). Finally, while model predictions are often made on the entity level, some use cases may require predictions on the word level, for instance if the associated probabilities are to be used for active learning.
Handling these technical intricacies requires expert knowledge.

\paragraph{(iii)}
There exists a multitude of NER-specific annotation schemes and variants and it is important to be aware of the differences. For instance, during data preprocessing, existing word or entity labels need to be mapped to token labels, which is an annotation scheme dependent process. At evaluation time, there are different ways to cope with predictions that do not obey the rules of the given annotation scheme (we will get back to this in Sec.~\ref{sec:advanced_usage_annotation_scheme}).

\paragraph{(iv)}
Training hyperparameters which lead to reasonable performance may depend on the employed model and dataset. For instance, while a small dataset often requires more training epochs, larger datasets can usually be trained for fewer epochs. \newline

The aim of \nerblackbox~is to provide a high-level framework which makes the usage of SOTA NER models as simple as possible. As we will see in detail in Sec.~\ref{sec:basic_usage}, it offers easy access to datasets from various sources, automated training and evaluation as well as simple but versatile model inference.
It does so by hiding all technical complications from the user\footnote{This is where the name \nerblackbox~stems from: The framework does not require any knowledge about internal processes and can be used as a black box by only specifying inputs (pretrained model, dataset) and using the outputs (fine-tuned model). Note that there is no direct relation to explainability.} and is targeted at developers as well as people who are not necessarily experts in machine learning or NLP.
However, \nerblackbox~also allows fine-grained control over all sorts of low-level parameters and provides many advanced features, some of which we will cover in Sec.~\ref{sec:advanced_usage}. This might make the library appealing also for researchers and experts.

\section{Related Work}
The most commonly used framework for transformer-based NLP is arguably the HuggingFace ecosystem, in particular the open source libraries \transformers, \datasets~and \evaluate. Another popular alternative is \spacy. 

High-level libraries that are build on top of \transformersNO~exist in the form of \simpletransformers~and \tner. 
\simpletransformersNO~is a high-level library that covers a broad range of NLP tasks with basic support for NER.
\tnerNO~is specific to NER with an emphasis on cross-domain and cross-lingual model evaluation. Of all the mentioned libraries, it is arguably the most similar to \nerblackbox. 
However, as will be discussed in the following sections, \nerblackbox~offers many unique and powerful features that---to the best of our knowledge---make it distinct from any existing frameworks.

\section{Basic Usage \label{sec:basic_usage}}

\begin{figure*}[!ht]
    \centering
    \includegraphics[scale=0.6]{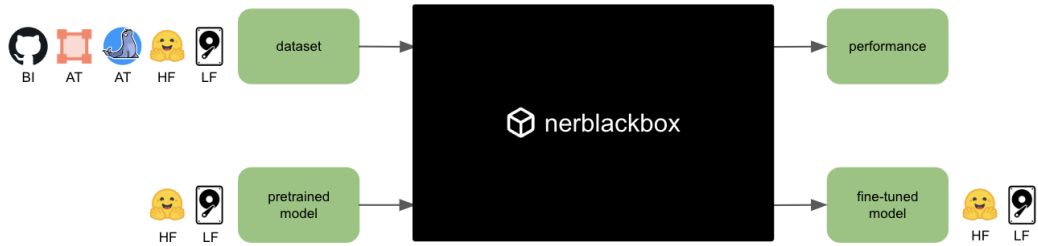}
    \caption{High-level overview of the \nerblackbox~library. It allows to easily fine-tune, evaluate and apply models for named entity recognition. The symbols to the left and right represent the sources that \nerblackbox~provides seamless access to. These are the Local Filesystem (LF), HuggingFace (HF), Annotation Tools (AT) as well as Built-in (BI) datasets that are fetched from GitHub.}
    \label{fig:nerblackbox_sources}
\end{figure*}

\nerblackbox~provides a simple API to automate each step in the life cycle of a NER model (cf.~Fig.~\ref{fig:workflow}) using very few lines of code.
It does so in terms of the following classes:
\begin{lstlisting}
>> from nerblackbox import Dataset, Training, Model
\end{lstlisting}

A high-level overview of the involved components is shown in Fig.~\ref{fig:nerblackbox_sources}.

\subsection{Dataset Integration \label{sec:basic_usage_dataset}}

\nerblackbox~allows seamless access to datasets from the following sources: HuggingFace (HF), the local filesystem (LF), built-in datasets (BI) and annotation tools (AT)\footnote{Currently, the two commonly used (open source) annotation tools \labelstudio~and \doccano~are supported.}.

Basically, a dataset can be set up for training and evaluation like in the following example:
\begin{lstlisting}
>> dataset = Dataset(
      "conll2003",  
      source="HF",
   )
>> dataset.set_up()
\end{lstlisting}

While this works out-of-the-box for the sources HF and BI, some additional information needs to be provided for the sources LF and AT in order for \nerblackbox~to be able to find the data.
Integrating different datasets can be challenging as they may have different formatting (even on HuggingFace) and annotation schemes. Some datasets are pretokenized and split into training/validation/test subsets, while others are not. 
The \code{set\_up()} method automatically deals with these challenges and makes sure that every dataset, irrespective of the source, is transformed into a standard format\footnote{Datasets may still have different annotation schemes (IO, BIO, BILOU), and be pretokenized or not.}. Apart from downloading, reformatting, and dataset splitting (if needed), it also includes an analysis of the data. For details, we refer to the library's documentation.

\subsection{Training \label{sec:basic_usage_training}}

In order to train a model, one only needs to choose a name for the training run (for later reference) and specify the model and dataset names, like so: 
\begin{lstlisting}
>> training = Training(
      "my_training", 
      model="bert-base-cased", 
      dataset="conll2003",
   )
>> training.run()
\end{lstlisting}

In order to ensure stable results irrespective of the dataset, the training employs well-established hyperparameters by default \cite{mosbach_stability_2021}. In particular, a specific learning rate schedule \cite{stollenwerk_adaptive_2022} based on early stopping and warm restarts \cite{loshchilov_sgdr_2017} is used to accommodate different dataset sizes.

\subsection{Evaluation \label{sec:basic_usage_evaluation}}

Any NER model, whether it was trained using \nerblackbox~or is taken directly from HuggingFace (HF), can be evaluated on any dataset that is accessible via \nerblackbox~(see Sec.~\ref{sec:basic_usage_dataset})
\begin{lstlisting}
>> model = Model.from_training(
      "my_training"
   )
>> results = model.evaluate_on_dataset(
      "conll2003", 
      phase="test",
   )
>> results["micro"]["entity"]["f1"]
## 0.9045
\end{lstlisting}

The standard metrics for NER are used, i.e. precision, recall and the f1 score. Each metric is computed as a micro- and macro-average as well as for the individual classes. All metrics are determined both on the entity and word level.

\subsection{Inference}
Similar to evaluation, both NER models trained using \nerblackbox~and models taken directly from HuggingFace (HF) can be used for inference. 

\begin{lstlisting}
>> model = Model.from_training(
      "my_training"
   )
>> model.predict("The United Nations")
## [[{
##   'char_start': '4', 
##   'char_end': '18', 
##   'token': 'United Nations', 
##   'tag': 'ORG'
## }]]
\end{lstlisting}

Apart from the predictions on the entity level for a single document shown above, \nerblackbox~also supports predictions on the word level (with or without probabilities) and batch inference. In addition, a model can be applied directly to a file containing raw data, which may be useful for inference at large scale (e.g. in production).

\section{Advanced Usage \label{sec:advanced_usage}}

The \nerblackbox~workflow and the API are designed to be as simple as possible and to conceal technical complications from the user. However, they are also highly customizable in terms of optional function arguments, which may be particularly interesting for machine learning experts and researchers. In this section, we are going to cover a non-exhaustive selection of \nerblackbox's advanced features, with a slight emphasis on the training part. For further information, the reader is referred to the library's documentation.

\subsection{Training Hyperparameters and Presets}

While \nerblackbox~uses sensible default values for the training hyperparameters (see Sec.~\ref{sec:basic_usage_training}), one may also opt to specify them manually. In particular, all aspects of the learning rate schedule (e.g. maximum learning rate, epochs, early stopping parameters etc.) can be chosen at one's own discretion. In addition, the \code{Training} class offers several popular hyperparameter presets via the instantiation argument \code{from\_preset}. Among them are the learning rate schedules from \cite{devlin_bert_2019} and \cite{mosbach_stability_2021}, which may work well for larger and smaller datasets, respectively.  Hyperparameters search is also supported.

\subsection{Dataset Pruning}

\nerblackbox~provides the option to only use a subset of the training, validation or test data by specifying parameters like \code{train\_fraction}. 
This may be useful to accelerate the training (for instance in the development phase of a product) or if one wants to investigate the effect of the dataset size (for instance to see if the model has saturated, or for research).

\subsection{Annotation Schemes}

While every dataset is associated with a certain annotation scheme, \nerblackbox~provides the option to translate between schemes at training time. The desired annotation scheme can simply be specified via the training parameter \code{annotation\_scheme}. This may be interesting for users who aim to optimize their model's performance as well as researchers who systematically want to investigate the impact of the annotation scheme.

\subsection{Multiple Runs}
Since the training of a neural network includes stochastic processes, the performance of the resulting model depends on the employed random seed. In order to gain control over the associated statistical uncertainties, one may train multiple models using different random seeds. With \nerblackbox, this can trivially be done by setting the training parameter \code{multiple\_runs} to an integer greater than 1. In that case, the evaluation metrics will be given in terms of the mean and its associated uncertainty. For inference, the best model is automatically used.

\subsection{Detailed Results}

\nerblackbox~saves detailed training and evaluation results (e.g. loss curves, confusion matrices) using \mlflow~and TensorBoard. This is useful in order to keep an overview of trained models, inspect their detailed properties as well as optimize and cross-check the training process.  

\subsection{Careful Evaluation \label{sec:advanced_usage_annotation_scheme}}

A model may predict labels for a sequence of tokens that are inconsistent with the employed annotation scheme. For instance, if the BIO annotation scheme is used, the combination \code{O I-PER} is incorrect\footnote{The variant of the BIO scheme which we assume here is also known as IOB2}. When translated to \textit{entity} predictions, \nerblackbox~ignores incorrect labels by default, both at evaluation and inference time. However, the popular \evaluate~and \seqeval~libraries do take inconsistent predictions into account during evaluation. For this reason, the \code{evaluate\_on\_dataset()} method (see Sec.~\ref{sec:basic_usage_evaluation}) returns results for both approaches.

\subsection{Compatibility with \transformersNO}

\nerblackbox~is heavily based on \transformers~such that compatibility is guaranteed. In particular, the \code{Model} class has the attributes \code{tokenizer} and \code{model}, which are ordinary \transformersNO~classes and can be used as such.
GPU support (i.e. automatic detection and use) is also provided through \transformersNO.

\section{Resources and Code Quality}

\nerblackbox~is available as a package on PyPI\footnote{\url{https://pypi.org/project/nerblackbox/}}. 
The associated GitHub repository is public at \url{https://github.com/flxst/nerblackbox} and contains the source code as well as multiple example notebooks.
A detailed documentation is provided\footnote{\url{https://flxst.github.io/nerblackbox/}}. It includes a pedagogical introduction to the library, an in-depth discussion of its features as well as docs for the python API.
Consistent code syntax and typing are ensured by usage of \black~and \mypy, respectively. We employ unit and end-to-end testing. As an additional cross-check, numerical results from the literature are reproduced using \nerblackbox~(details can be found in the documentation).

\section*{Acknowledgements}

This work was supported by Vinnova through the grants 2019-02996 and 2021-03630.

\bibliography{anthology,NER}
\bibliographystyle{acl_natbib}

\end{document}